\documentclass[times]{mcp-acm}
\pdfoutput=1

\usepackage{amsmath}    
\usepackage{graphicx}   

\makeindex              

\usepackage{url}
\usepackage{amsfonts}
\usepackage{amssymb}
\usepackage{subfigure}
\usepackage{comment}
\usepackage{epsfig}
\usepackage{xspace}
\usepackage{epstopdf}

\usepackage{color}

\usepackage{enumitem}
\usepackage{algorithm}
\usepackage{algorithmic}
\usepackage{multirow}
\usepackage{balance}
\usepackage{booktabs}
\usepackage{array}
\usepackage{wrapfig}
\usepackage{pifont}

\usepackage{etoolbox}

\DeclareGraphicsExtensions{.pdf,.eps,.png,.jpg,.tif,.tiff,.ps}

\usepackage{geometry}
\usepackage{bm}
\usepackage{latexsym}
\usepackage{color}
\usepackage{setspace}
\usepackage{mathrsfs}
\usepackage{graphicx}
\usepackage{listings}
\usepackage{scrextend}
\usepackage{enumitem}
\usepackage{subfigure}
\usepackage[utf8]{inputenc}

\usepackage{tikz}
\usetikzlibrary{decorations.markings}
\usepackage{times}
\usepackage{epstopdf}
\usepackage{hyperref}
\usepackage{amsmath}
\usepackage{amssymb}
\usepackage{multirow}
\usepackage{siunitx}
\usepackage{booktabs}
\usepackage{arydshln}

\usepackage{bbm}

\usepackage{pgfplots}

\def\MarkLt{6pt}
\def\MarkSep{3pt}
\tikzset{
	TwoMarks/.style={
		postaction={decorate,
			decoration={
				markings,
				mark=at position #1 with
				{
					\begin{scope}[xslant=0.2]
						\draw[line width=\MarkSep,white,-] (0pt,-\MarkLt) -- (0pt,\MarkLt) ;
						\draw[-] (-0.5*\MarkSep,-\MarkLt) -- (-0.5*\MarkSep,\MarkLt) ;
						\draw[-] (0.5*\MarkSep,-\MarkLt) -- (0.5*\MarkSep,\MarkLt) ;
					\end{scope}
				}
			}
		}
	},
	TwoMarks/.default={0.5},
	OneMark/.style={
		postaction={decorate,
			decoration={
				markings,
				mark=at position #1 with
				{
					\draw[-] (0,-\MarkLt) -- (0,\MarkLt) ;
				}
			}
		}
	},
	OneMark/.default={0.5}
}

\makeatletter
\newcommand{\vast}{\bBigg@{3}}
\newcommand{\Vast}{\bBigg@{4}}
\makeatother

\definecolor{navy}{rgb}{0.1, 0.1, 0.8}
\definecolor{gray}{rgb}{0.4, 0.4, 0.4}
\definecolor{ruby}{rgb}{0.8, 0.1, 0.1}

\newcommand{\eat}[1]{}

\usepackage{soul}

\input{macros-common}

\begin{document}

\frontmatter

\mainmatter



\chapter{A Tutorial on Hawkes Processes for Events in Social Media}

Marian-Andrei Rizoiu, The Australian National University; Data61, CSIRO\\
Young Lee, Data61, CSIRO; The Australian National University\\
Swapnil Mishra, The Australian National University; Data61, CSIRO\\
Lexing Xie, The Australian National University; Data61, CSIRO\\\\
	
\newtheorem{theorem}{Theorem}
\newtheorem{definition}{Definition}
\newtheorem{remark}{Remark}
\newtheorem{proposition}{Proposition}
\newtheorem{corollary}{Corollary}
\newtheorem{lemma}{Lemma}
\newcommand{\p}{\mathbb{P}}
\newcommand{\q}{\mathbb{Q}}
\newcommand{\e}{\mathbb{E}}
\newcommand{\setar}{\textcolor{blue}{$\star\star\star\quad$}}

\definecolor{keywords}{RGB}{255,0,90}
\definecolor{comments}{RGB}{0,0,113}
\definecolor{red}{RGB}{160,0,0}
\definecolor{green}{RGB}{0,150,0}
	
\lstset{language=Python, 
		basicstyle=\ttfamily\small, 
		keywordstyle=\color{keywords},
		commentstyle=\color{comments},
		stringstyle=\color{red},
		showstringspaces=false,
		identifierstyle=\color{green}}

This chapter provides an accessible introduction for point processes, and especially Hawkes processes, for modeling discrete, inter-dependent events over continuous time.
We start by reviewing the definitions and the key concepts in point processes. 
We then introduce the Hawkes process, its event intensity function, as well as schemes for event simulation and parameter estimation. 
We also describe a practical example drawn from social media data -- we show how to model retweet cascades  using a Hawkes self-exciting process. We presents a design of the memory kernel, and results on estimating parameters and predicting popularity. The code and sample event data are available as an online appendix. 
	
	


\section{Introduction}
\label{sec:intro}

\begin{figure}[b]
	\centering
	\includegraphics[width=0.99\textwidth]{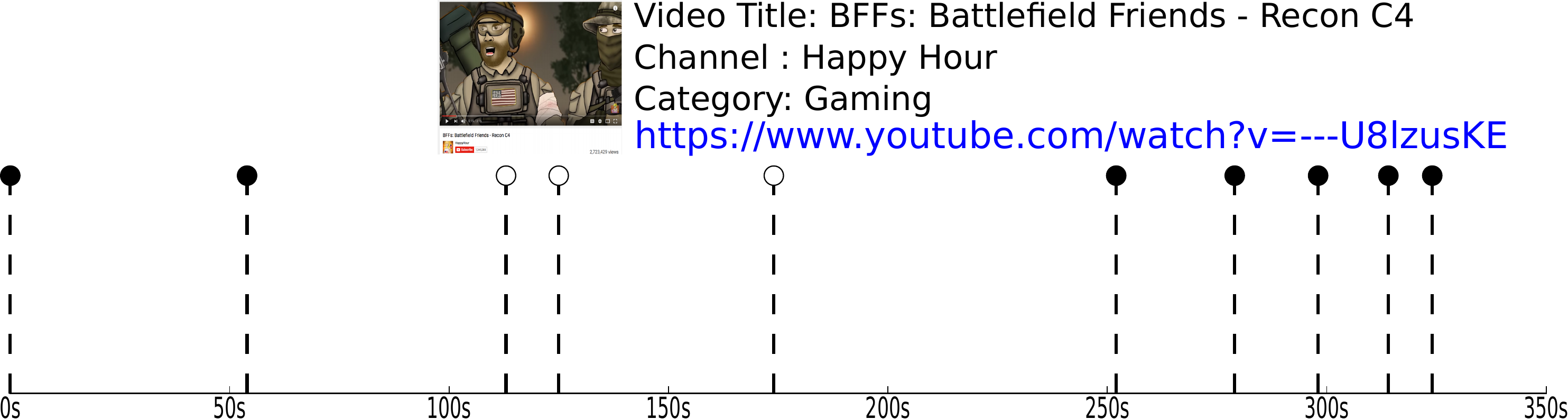}
	\caption{An point process, showing tweets about \href{https://www.youtube.com/watch?v=---U8lzusKE}{a Gaming video} on Youtube. 
	The first 10 events are shown. They correspond to the first 10 tweets in the diffusion, the time stamps of which are indicated by dashed vertical lines.
	An event with hollow tip denote a retweet of a previous tweet.
	}
	\label{fig:pointprocess}
\end{figure}

Point processes are collections of random points falling in some space, such as time and location.
Point processes provide the statistical language to describe the timing and properties of events. 
Problems that fit this setting span a range of application domains. 
In finance, an event can represent a buy or a sell transaction on the stock market 
that influences future prices and volumes of such transactions.
In geophysics, an event can be an earthquake that is indicative of the likelihood of another earthquake in the vicinity in the immediate future.
In ecology, event data consist of a set of point locations where a species has been observed. 
In the analysis of online social media, events can be user actions over time, each of which have a set of properties such as user influence, topic of interest, and connectivity of the surrounding network.

Fig.~\ref{fig:pointprocess} depicts an example point process -- a retweet cascade about a Gaming Youtube video (YoutubeID \emph{-{}-{}-U8lzusKE}).
Here each tweet is an event, that happens at a certain point in continuous time.
Three of the events depicted in Fig.~\ref{fig:pointprocess} are depicted using hollow tips -- they are retweets of a previous tweet, or the act of one user re-sharing the content from another user.
We explicitly observe information diffusion via retweets, however there are other diffusion mechanisms that are not easily observed. These include offline \emph{word-of-mouth} diffusion, or information propagating in emails and other online platforms.
One way for modeling the overall information diffusion process is to use so called {\em self-exciting processes} -- in this type of processes the probability of seeing a new event increases due to previous events.
Point-process models are useful for answering a range of different questions. These include {\em explaining} the nature of the underlying process, {\em simulating} future events, and {\em predicting} the likelihood and volume of future events.

In Section~\ref{sec:point-poisson-processes}, we first review the basic concepts and properties of point processes in general, and of Poisson processes. 
These provide foundations for defining the Hawkes process.
In Section~\ref{sec:hawkes-processes}, we introduce the Hawkes process -- including expressions of the event rate and the underlying branching structure.
Section~\ref{sec:simulation} describes two procedures for sampling from a Hawkes process. One uses thinning, rejection sampling, while the other make use of a novel variable decomposition.
Section~\ref{sec:fitting-data} derives the likelihood of a Hawkes process 
and describes a maximum likelihood estimation procedure for its parameters, given observed event sequence(s).
In the last section of this chapter we present an example of estimating Hawkes processes from a retweet event sequence. 
We introduce the data, the problem formulation, the fitting results, and interpretations of the model. 
We include code snippets for this example in Sec~\ref{ssec:handson}, the accompanying software and data are included in an \href{https://github.com/s-mishra/featuredriven-hawkes}{online repository}.

This chapter aims to provide a self-contained introduction to the fundamental concepts and methods for self-exciting point-processes, with a particular emphasis on the Hawkes process. 
The goal is for the readers to be able to understand the key mathematical and computational constructs of a point process, formulate their problems in the language of point processes, and use point process in domains including but are not limited to modeling events in social media. 
The study of point processes has a long history, with discussions of Hawkes process dating back at least to the early 1970s~\citep{hawkes71}.
Despite the richness of existing literature, we found through our own recent experience in learning and applying Hawkes processes, that a self-contained tutorial centered around problem formulation and applications is still missing.
This chapter aims at filling this gap, providing the foundations as well as an example of point processes for social media. Its intended audiences are aspiring researchers and beginning PhD students, as well as any technical readers with a special interest in point processes and their practical applications.
For in-depth reading, we refer the readers to overview papers and books~\citep{DaleyBook,toke2011introduction} on Hawkes processes. 
We note that this chapter does not cover other important variants used in the multimedia area, such as self-inhibiting processes~\cite{Yang2015}, or non-causal processes (in time or space), such as the Markov point processes~\cite{pham2016efficient}.

%
%
%
%
%


\section{Preliminary: Poisson processes}
\label{sec:point-poisson-processes}

In this section, we introduce
the fundamentals of point processes and its simplest subclass, the Poisson process.
These serve as the foundation on which we build, in later sections, the more complex processes, such as the Hawkes point process.

\subsection{Defining a point process}

A point process on the nonnegative real line, where the nonnegative line is taken to represent time, is a random process whose realizations consists of the event times $T_1,T_2, \ldots$ of event times falling along the line. 
$T_i$ can usually be interpreted as the time of occurrence of the $i$-th event, and $T_i$ are often referred to as event times.
	
\textbf{The equivalent counting process.}
A counting process $N_t$ is a random function defined on time $t\geq 0$, and take integer values $1,2,\ldots$. 
Its value is the number of events of the point process by time $t$. Therefore it is uniquely determined by a sequence of non-negative random variables $T_i$, satisfying $T_i<T_{i+1}$ if $T_i\leq\infty$. In other words, $N_t$ {\em counts} the number of events up to time $t$, i.e.
\begin{align}
	{N}_t:=\sum_{i\geq 1}\mathbbm{1}_{\{t\geq T_i\}}
\end{align}
Here $\mathbbm{1}_{\{\cdot\}}$ is the indicator function that takes value 1 when the condition is true, 0 otherwise. We can see that $N_0=0.$ $N_t$ is piecewise constant and has jump size of 1 at the event times $T_i$. 
It is easy to see that the set of event times $T_1,T_2, \ldots$ and the corresponding counting process are equivalent representations of the underlying point process. 
	
\subsection{Poisson processes: definition}
The simplest class of point process is the Poisson process. 

\begin{definition} \label{def:poisson-process}
	(Poisson process.)
	Let $(\tau_i)_{i\geq 1}$ be a sequence of \textit{i.i.d.} exponential random variables with parameter $\lambda$ and event times $T_n=\sum_{i=1}^{n}\tau_i$. 
	The process $(N_t,t\geq 0)$ defined by ${N}_t:=\sum_{i\geq 1}\mathbbm{1}_{\{t\geq T_i\}}$ is called a \textit{Poisson process} with intensity $\lambda$.
\end{definition}

\textbf{Event intensity $\lambda$.}
The sequence of $\tau_j$ are called the \emph{inter-arrival times}, i.e. the first event occurs at time $\tau_1$, the second occurs at $\tau_2$ after the first, etc.
The inter-arrival times $\tau_i$ are independent, and each of them follow an exponential distribution with parameter $\lambda$. Here, the notation $f_{\tau} (t)$ denotes the probability density function of 
random variable $\tau$ taking values denoted by $t$. 

\begin{align}
	f_\tau (t)= 
	\begin{cases}
	\lambda e^{-\lambda t},& \text{if } t\geq 0\\
	0,              & \text{if } t< 0
	\end{cases}
\end{align}

Here $\lambda>0$ is a positive constant. 
The expected value of $\tau_i$ can be computed in closed form, as follows:

\begin{align} \label{eq:pdf-exp}
	\e_\tau[\tau]&=\int_0^{\infty}t f_{\tau}(t)dt = \lambda\int_0^{\infty}t e^{-\lambda t}dt = \left[-t e^{-\lambda t}\right]_{t=0}^{t=+\infty}+\int_0^{\infty}e^{-\lambda t} dt \nonumber\\
	&=0-\left[\frac{1}{\lambda}e^{-\lambda t}\right]_{t=0}^{t=\infty}=\frac{1}{\lambda}.	
\end{align}

Intuitively, events are arriving at an average rate of $\lambda$ per unit time, since the expected time between event times is $\lambda^{-1}$. 
Hence we say, informally, that the Poisson process has \textit{intensity} $\lambda$.
In general, the event intensity needs not be constant, but is a function of time, written as $\lambda(t)$. This general case is called a \emph{non-homogeneous Poisson process}, and will be discussed in Sec.~\ref{subsec:non-homogeneous-poisson}.

\textbf{Arrival times and counting process.}
The \textit{arrival times}, or the event times, are given by:
\begin{align}
	T_n=\sum_{j=1}^n \tau_j,
\end{align} 
where $T_n$ is the time of the $n$-th arrival. 
The event times $T_1,T_2,...$ form a random configuration of points on the real line $[0,\infty)$ and $N_t$ counts the number of such ons in the interval $[0,t]$.
Consequently, $N_t$ increments by one for each $T_i$. This can be explicitly written as follows. 
\begin{equation} \label{eq:Nt-detailed}
	N_t= 
	\begin{cases}
	0,& \text{if } 0\leq t < T_1 \\
	1,& \text{if } T_1\leq t < T_2 \\
	2,& \text{if } T_2\leq t < T_3 \\
	\vdots\\
	n,& \text{if } T_n\leq t < T_{n+1}, \\
	\vdots\\
	\end{cases}
\end{equation}
We observe that $N_t$ is defined so that it is 
\emph{right continuous with left limits}. 
The left limit $N_{t-}=\lim_{s\uparrow t}N_s$ exists and $N_{t+}=\lim_{s\downarrow t}N_s$ exists and taken to be $N_t$. 


\subsection{The memorylessness property of Poisson processes}


Being {\em memoryless} in a point process means that the distribution of future inter-arrival times depends only on relevant information about the current time, but not on information from further in the past. 
We show that this is the case for Poisson processes. 

We compute the probability of observing an inter-arrival time $\tau$ longer than a predefined time length $t$.
$F_{\tau}$ is the cumulative distribution function of the random variable $\tau$, which is defined as $F_{\tau}(t):=\p\{\tau\leq t\}$. 
We have
\begin{align}
	F_\tau(t):=\p(\tau\leq t)=\int_0^t \lambda e^{-\lambda x}dx=\left[-e^{\lambda x}\right]_{x=0}^{x=t}=1-e^{-\lambda t},\quad t\geq 0,
\end{align}
and hence the probability of observing an event at time $\tau>t$ is given by
\begin{align}
	\p(\tau>t)=e^{-\lambda t},\quad t\geq 0.\label{eq:poisson_time}
\end{align}
	
Suppose we were waiting for an arrival of an event, say a tweet, the inter-arrival times of which follow an Exponential distribution with parameter $\lambda$. 
Assume that $m$ time units have elapsed and during this period no events have arrived, i.e. there are no events during the time interval $[0,m]$. 
The probability that we will have to wait a further $t$ time units given by
\begin{align} \label{eq:memorylessness}
	\p(\tau>t+m|\tau>m)&=\frac{\p(\tau>t+m,\tau>m)}{\p(\tau>m)} \nonumber \\
	&=\frac{\p(\tau>t+m)}{\p(\tau>m)}=\frac{e^{-\lambda(t+m)}}{e^{-\lambda m}}=e^{-\lambda t}=\p(\tau>t).
\end{align}
In this derivation, we first expand the conditional probability using Bayes rule. The next step follows from the fact that $\tau>m$ always holds when $\tau>t+m$. The last step follows from Eq.~\eqref{eq:poisson_time}.

Eq.~\eqref{eq:memorylessness} denotes the {\em memorylessness} property of Poisson processes. That is, the probability of having to wait an additional $t$ time units after already having waited $m$ time units is the same as the probability of having to wait $t$ time units when starting at time $0$. 
Putting it differently, if one interprets $\tau$ as the time of arrival of an event where $\tau$ follows an Exponential distribution, the distribution of $\tau-m$ given $\tau>m$ is the same as the distribution of $\tau$ itself.

\subsection{Non-homogeneous Poisson processes}
\label{subsec:non-homogeneous-poisson}
	
In Poisson processes, events arrive randomly with the constant intensity $\lambda$.
This initial model is sufficient for describing simple processes, say the arrival of cars on a street over a short period of time. 
However, we need to able to vary the event intensity with time in order to describe more complex processes, such as simulating the arrivals of cars during rush hours and off-peak times.
In a non-homogeneous Poisson process, the rate of event arrivals is a function of time, i.e. $\lambda = \lambda(t)$.
\begin{definition}
	A point process $\{N_t\}_{t>0}$ can be completely characterized by its conditional intensity function, defined as
	\begin{align}
		\lambda(t|\mathcal{H}_t)=\lim_{h\rightarrow 0}\frac{\p\{N_{t+h}-N_t=1|\mathcal{H}_t\}}{h}
	\end{align}
	where $\mathcal{H}_t$ is the history of the process up to time $t$, containing the list of event times $\{T_1,T_2,...,T_{N_t}\}$. 
\end{definition}
In the rest of this chapter, we use the shorthand notation $\lambda(t)=:\lambda(t|\mathcal{H}_t)$, always assuming an implicit history before time $t$.
The above definition gives the intensity view of a point process, equivalent with the two previously defined views with events times and the counting process. 
In other words, the event intensity $\lambda(t)$ determines the distribution of event times, which in turn determine the counting process.
Formally, $\lambda(t)$ and $N_t$ are related through the probability of an event in a small time interval $h$: 
\begin{align} \label{eq:non-homog-poisson}
	\p(N_{t+h}=n+m\,|\,N_t=n)&=\lambda(t) h+o(h)\qquad&\mathrm{if}\qquad m=1 \nonumber\\
	\p(N_{t+h}=n+m\,|\,N_t=n)&=o(h)&\qquad\mathrm{if}\qquad m>1\nonumber\\
	\p(N_{t+h}=n+m\,|\,N_t=n)&=1-\lambda(t) h + o(h)&\qquad\mathrm{if}\qquad m=0 
\end{align}
where $o(h)$ is a function so that $\lim_{h \downarrow 0}\frac{o(h)}{h} = 0$.
In other words, the probability of observing an event during the infinitesimal interval of time $t$ and $t +h$ when $h \downarrow 0$ is $\lambda(t) h$.
The probability of observing more than one event during the same interval is negligible.



\section{Hawkes processes}
\label{sec:hawkes-processes}

In the models described in the previous section, the events arrive independently, either at a constant rate (for the Poisson process) or governed by an intensity function (for the non-homogeneous Poisson).
However, for some applications, it is known that the arrival of an event increases the likelihood of observing events in the near future.
This is the case of earthquake aftershocks when modeling seismicity, or that of user interactions when modeling preferential attachment in social networks.
In this section, we introduce a class of processes in which the event arrival rate explicitly depends on past events -- i.e. \emph{self-exciting processes} -- and we further detail the most well-known self-exciting process, the Hawkes process.

\subsection{Self-exciting processes}

A self-exciting process is a point process in which the arrival of an event causes the conditional intensity function to increase.
A well known self-exciting process was proposed by \citet{hawkes71}, and it is based on a counting process in which the intensity function depends explicitly an all previously occurred events. The Hawkes process is defined as follows: 

\begin{definition} (Hawkes process)
	Let $\{N_t\}_{t>0}$ be a counting process with associated history $\mathcal{H}_t,t\geq 0$.
	The point process is defined by the event intensity function $\lambda(t)$ with respects Eq.~\eqref{eq:non-homog-poisson} (the intensity view of a non-homogeneous Poisson process).
	The point process is said to be a Hawkes process if the conditional intensity function $\lambda(t|\mathcal{H}_t)$ takes the form:
	\begin{align} \label{first-lambda}
		\lambda(t|\mathcal{H}_t) = \lambda_0(t)+\sum_{i:t>T_i}\phi(t-T_i) \enspace,
	\end{align}
\end{definition}
where $T_i < t$ are all the event time having occurred before current time $t$, and which contribute to the event intensity at time $t$.
$\lambda_0(t):\mathbb{R}\mapsto\mathbb{R}_+$ is a deterministic base intensity function, 
and $\phi:\mathbb{R}\mapsto\mathbb{R}_+$ is called the memory kernel -- both of which are further detailed in the next section.
We observe that the Hawkes process is a particular case of non-homogeneous Poisson process, in which the intensity is stochastic and explicitly depends on previous events through the kernel function $\phi(\cdot)$.

\begin{figure}[htbp]
	\centering
	
    \includegraphics[width=0.95\textwidth]{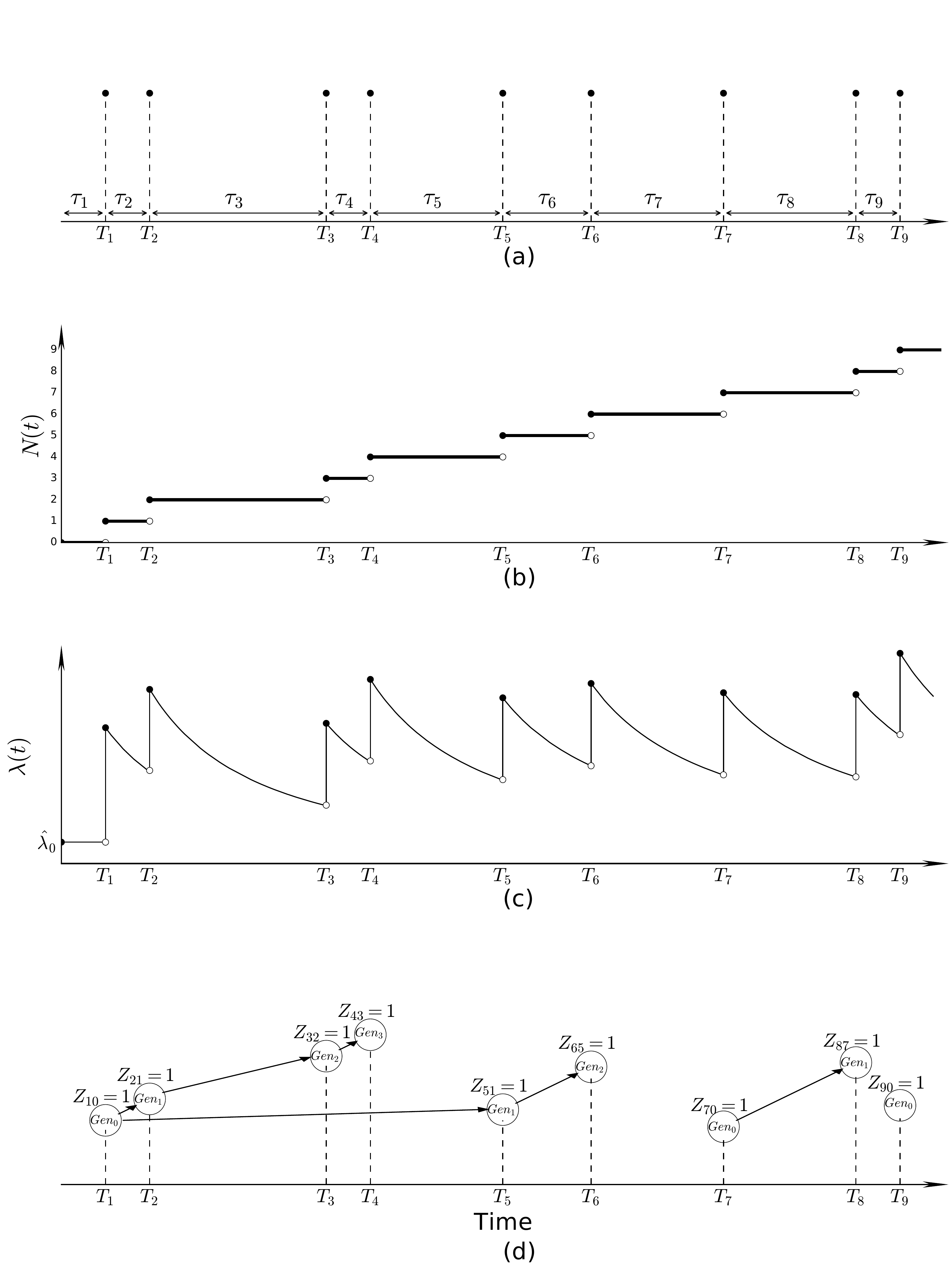}
      
	\caption{Hawkes process with an exponential decay kernel.
	(a) The first nine event times are shown. 
	$T_i$ represent event times, while $\tau_i$ represent inter-arrival times.
	(b) Counting process over time, $N_t$ increases by one unit at each event time $T_i$.
	(c) Intensity function over time.
	Note how each event provokes a jump, followed by an exponential decay.
	Later decays unfold on top of the tail of earlier decays, resulting in apparently different decay rates.
	(d) The latent or unobserved branching structure of the Hawkes process.
	Every circle represents one event having occurred at $T_i$, the arrows represent the root-offspring relation.
	$\mathcal{G}en_i$ specifies the generation of the event, with $i=0$ for immigrants or $i>0$ for the offspring. $Z_{ij}$ are random variables, such that $Z_{i0}=1$ if event $i$ is an immigrant, and $Z_{ij}=1$ if event $i$ is an offspring of event $j$.
	}
	\label{fig:hawkes-process-complete}
\end{figure}

\subsection{The intensity function}
\label{ssec:intensity_func}
The quantity $\lambda_0(t)>0$ is the base (or background) intensity, describing the arrival of events triggered by external sources.
These events are also known as \textit{exogenous} or {\em immigrant} events, and their arrival is independent on the previous events within the process.
The self-exciting flavor of the Hawkes process arises through the summation term in Eq.~\eqref{first-lambda}, where 
the kernel $\phi(t - T_i)$ modulates the change that an event at time $T_i$ has on the intensity function at time $t$.
Typically, the function $\phi(\cdot)$ is taken to be monotonically decreasing so that more recent events have 
higher influence on the current event intensity, compared to events having occurred further away in time.
Fig.~\ref{fig:hawkes-process-complete}(a) shows an example realization of a Hawkes process: nine events are observed, at times $T_1, T_2, \dots, T_9$, and their corresponding inter-arrival times $\tau_1, \tau_2, \cdots, \tau_9$.
Fig~\ref{fig:hawkes-process-complete}(b) shows the corresponding counting process $N_t$ over time, which increases by one unit for each $T_i$ as defined in Eq.~\eqref{eq:Nt-detailed}.
Fig.~\ref{fig:hawkes-process-complete}(c) shows the intensity function $\lambda(t)$ over time.
Visibly, the value of the intensity function increases suddenly immediately at the occurrence of an event $T_i$, and diminishes as time passes and the effect of the given event $T_i$ decays. 
	
\textbf{Choice of the kernel $\phi$.}	
The kernel function $\phi(\cdot)$ does not have to be monotonically decreasing.
However, in this chapter we restrict the discussion to the decreasing families of functions, given that it is natural to see the influence of an event decay over time, as shown in Sec.~\ref{sec:hawkes-social-media}.
A popular decay function is the exponential function~\citep{hawkes71}, taking the following form:
\begin{equation} \label{eq:exp-kernel}
	\phi(x)=\alpha e^{-\delta x},
\end{equation}
where $\alpha\geq 0,\,\delta>0$ and $\alpha<\delta$. 
Another kernel that is widely used in the literature is the power-law kernel:
\begin{equation}
	\phi(x)=\frac{\alpha}{(x+\delta)^{\eta+1}},
\end{equation}
where $\alpha\geq 0,\,\delta,\eta>0$ and $\alpha<\eta\delta^{\eta}$. 
This kernel is commonly used within the seismology literature \citep{Ozaki1979} and in the social media literature~\citep{Rizoiu2017}. 
The exponential kernel defined by Eq.~\ref{eq:exp-kernel} is typically the popular choice of kernel with Hawkes processes~\citep{Liniger2011}, unless demanded otherwise by the phenomena modeled using the self-exciting process (for example, we use a power-law kernel for modeling information diffusion in Social Media, in Sec.~\ref{sec:hawkes-social-media}).
	
\textbf{Other self-exciting point processes}
have been proposed, which follow the canonical specification given in Eq.~\eqref{first-lambda} and which extend the initial self-exciting process proposed by~\citet{hawkes71}.
We do not cover these processes in this chapter, however we advise the reader of Hawkes extensions
such as the non-linear Hawkes processes~\citep{BremaudMassoulie1996,DaleyBook}, 
the general space time self-exciting point process~\citep{veen-schoenberg,Ogata1988},
processes with exponential base event intensity~\citep{biao-aap}, or self-inhibiting processes~\citep{Yang2015}. 
	
\subsection{The branching structure}
\label{subsec:branching-structure}

Another equivalent view of the Hawkes process refers to the Poisson cluster process interpretation~\citep{hawkes-oakes-cluster}, which separates the events in a Hawkes process into two categories: \textit{immigrants} and \textit{offspring}. 
The offspring events are tiggered by existing (previous) events in the process, while the immigrants arrive independently and thus do not have an existing parent event.
The offspring are said to be structured into \emph{clusters}, associated with each immigrant event.
This is called \emph{the branching structure}.
In the rest of this section, we further details the branching structure and we compute two quantities:
the \emph{branching factor} -- the expected number of events directly triggered by a given event in a Hawkes process -- and the estimated total number of events in a cluster of offspring.
As shown in Sec.~\ref{sec:hawkes-social-media}, both of these quantities become very important when the Hawkes processes are applied to practical domains, such as online social media.

\textbf{An example branching structure.}
We consider the case that immigrant events follow a homogeneous Poisson process with base intensity  $\lambda_0(t)$, while offspring are generated through the self-excitement, governed by the summation term in Eq.~\eqref{eq:hawkes}.
Fig.~\ref{fig:hawkes-process-complete}(d) illustrates the branching structure of the nine event times of the example Hawkes process discussed earlier.
Event times $T_i$ are denoted by circles and the `parent-offspring' relations between the events are shown by arrows.
We introduce the random variables $Z_{ij}$, where $Z_{i0}=1$ if event $i$ is an immigrant, and $Z_{ij}=1$ if event $i$ is an offspring of event $j$. 
The text in each circle denotes the generation to which the event belongs to, i.e. $\mathcal{{G}}en_k$ denotes the \textit{k}-th generation.
Immigrants are labeled as $\mathcal{{G}}en_0$, while generations $\mathcal{{G}}en_k, \, k > 0$ denote their offspring.
%
For example $T_3$ and $T_6$ are immediate offspring of the immigrant $T_2$, i.e. mathematically expressible as $Z_{32} = 1$, $Z_{62} = 1$ and $Z_{20} = 1$.

The cluster representation states that the immediate offspring events associated with a particular parent arrive according to a non-homogeneous Poisson process with intensity $\phi(\cdot)$, i.e. $T_3$ and $T_6$ are event realizations coming from a non-homogeneous Poisson process endowed with intensity $\phi(t-T_2)$ for $t>T_2$. 
The event that produces an offspring is described as the immediate ancestor or root of the offspring, $T_7$ is the immediate ancestor of $T_8$. 
The events which are directly or indirectly connected to an immigrant form the \textit{cluster} of offspring associated with that immigrant, e.g. $T_1$ is an immigrant and $T_2,T_3,T_4,T_5$ and $T_6$ form its cluster of offspring. 
Similarly, $T_7$ and $T_8$ form another cluster. 
Finally, $T_9$ is a cluster by itself.

\textbf{Branching factor} (branching ratio).
One key quantity that describes the Hawkes processes is its branching factor $n^{\ast}$, defined as the expected number direct offspring spawned by a single event.
The branching factor $n^{\ast}$ intuitively describes the amount of events to appear in the process, or informally, {\em virality} in the social media context.
In addition, the branching factor gives an indication about whether the cluster of offspring associated with an immigrant is an infinite set.
For $n^{\ast} < 1$, the process in a \emph{subcritical regime}: the total number of events in any cluster is bounded. 
Immigrant event occur according to the base intensity $\lambda_0(t)$, but each one of them has associated with it a finite cluster of offspring, both in number and time extent.
When $n^{\ast} > 1$, the process is in a so-called \emph{supercritical regime} with $\lambda(t)$ increasing and the total number of events in each cluster being unbounded.
We compute the branching factor by integrating $\phi(t)$ -- the contribution of each event -- over event time $t$:
\begin{equation} \label{eq:nstar-general}
	n^{\ast} = \int_{0}^\infty \phi(\tau) d\tau \enspace.
\end{equation}

\textbf{Expected number of events in a cluster of offspring.}
The branching factor $n^*$ indicates whether the number of offspring associated with each immigrant is finite ($n^* < 1$) or infinite ($n^* > 1$).
When $n^* < 1$ a more accurate estimate of the size of each cluster can be obtained.
Let $A_i$ be the expected number of events in $Generation_i$, and $A_0 = 1$ (as each cluster has only one immigrant).
The expected number of total events in the cluster, $N_{\infty}$, is defined as:
\begin{equation}\label{eq:n-inf-cluster}
	N_{\infty} = \sum_{i=0}^{\infty} A_i \enspace.
\end{equation}
To compute $A_i, i \ge 1$, we notice that each of the $A_{i-1}$ events in the previous generation has on average $n^{\ast}$ children events. This leads to a inductive relationship $A_i = A_{i-1} n^*$.
Knowing that $A_0 = 1$, we derive:
\begin{equation} \label{eq:gp}
	A_i = A_{i-1} \: n^\ast = A_{i-2} \: \left( n^\ast \right)^2 = \ldots\ = A_0 \:  \left( n^\ast \right)^{i}  = \left( n^\ast \right)^{i}, i \geq 1
\end{equation}
We obtain an estimate of the size of each cluster of immigrants $N_{\infty}$ as the sum of a converging geometric progression (assuming $n^{\ast} < 1$):
\begin{equation} \label{eq:n-inf-cluster-sumgp}
	N_{\infty} = \sum_{i=0}^{\infty} A_i = \dfrac{1}{1-n^\ast} \textrm{ where } n^\ast < 1
\end{equation}

\section{Simulating events from Hawkes processes}
\label{sec:simulation}

In this section, we focus on the problem of simulating series of random events according to the specifications of a given Hawkes process.
This is a useful for gathering statistics about the process, and can form the basis for diagnostics, inference or parameter estimation. 
%
We present two simulation techniques for Hawkes processes.
The first technique, the thinning algorithm~\citep{Ogata1981}, applies to all non-homogeneous Poisson processes, and can be applied to Hawkes processes with any kernel function $\phi(\cdot)$.
The second technique, recently proposed by \citet{biao-ecp}, is computationally more efficient, as it designs a variable decomposition technique for Hawkes processes with exponential decaying kernels.

\subsection{The thinning algorithm}
\label{subsec:thinning-simulation}

The basic goal of an sampling algorithm is to simulate inter-arrival times $\tau_i$, $i=1,2,\dots$ according to an intensity function $\lambda_t$. 
We first review the sampling method for a homogeneous Poisson process, then we introduce the thinning (or additive) property of Poisson processes, and we use this to derive the sampling algorithm for Hawkes processes. 

Inter-arrival times in a homogeneous Poisson process follow an exponential distribution as specified in~\ref{eq:pdf-exp}:
$f_{\tau}(t) = \lambda e^{-\lambda t},~t>0$
and its cumulative distribution function is $F_{\tau}(t) = 1- e^{-\lambda t}$. 
Because both $F_{\tau}(t)$ and $F^{-1}_{\tau}(t)$ have a closed-form expression, we can use the \emph{inverse transform sampling} technique to sample waiting times.
Intuitively, if $X$ is a random variable with the cumulative distribution function $F_X$ and $Y = F_X(X)$ is a uniformly distributed random variable ($\sim U(0, 1)$), then $X^* = F^{-1}_X(Y)$ has the same distribution as $X$.
In other words, sampling $X^* = F^{-1}_X(Y), Y \sim U(0, 1)$ is identical with sampling $X$.
For the exponentially distributed waiting times of the Poisson process, the inverse cumulative distribution function has the form
$F^{-1}_{\tau}(u) = \frac{-\ln u}{\lambda} $.
Consequently, sampling a waiting interval $\tau$ in a Poisson process is simply:
\begin{equation}
	\text{Sample } u\sim U(0, 1),~ \text{then compute } \tau = \frac{-\ln u}{\lambda}			\label{eq:sample_poisson}
\end{equation}

The thinning property of the Poisson processes states that a Poisson process with the intensity $\lambda$ can be split into two independent processes with intensities $\lambda_1$ and $\lambda_2$, so that $\lambda = \lambda_1 + \lambda_2$. In other words, each event of the original process can be assigned to one of the two new processes that are running independently.
From this property, we can see that we can simulate of a non-homogeneous Poisson process with the intensity function $\lambda(t)$ 
by \emph{thinning} a homogeneous Poisson process with the intensity $\lambda^* \geq \lambda(t), \forall t$.

A thinning algorithm to simulate Hawkes processes is presented in Algorithm~\ref{alg:simulation-thinning}. 
For any bounded $\lambda(t)$ we can find a constant $\lambda^*$ so that $\lambda(t) \leq \lambda^*$ in a given time interval.
In particular, for Hawkes processes with a monotonically decreasing kernel function $\phi(t)$, it is easy to see that between two consecutive event times $[T_i, T_{i+1})$, $\lambda(T_i)$ is the upper bound of event intensity. 
We exemplify the sampling of event time $T_{i+1}$, after having already sampled $T_1, T_2, \ldots, T_i$.
We start our time counter $T = T_i$.
We sample an inter-arrival time $\tau$, using Eq.~\eqref{eq:sample_poisson}, with $\lambda^* = \lambda(T)$ and we update the time counter $T = T + \tau$ (steps \ref{step:step-3a} to \ref{step:step-3c} in Algorithm~\ref{alg:simulation-thinning})
We accept or reject this inter-arrival time according to the ratio of the true event rate to the thinning rate $\lambda^*$ (step \ref{step:step-3e}).
If accepted, we record the event time $i+1$ as $T_{i+1} = T$.
Otherwise, we repeat the sampling of an inter arrival time until one is accepted.
Note that, even if an inter-arrival time is rejected, the time counter $T$ is still updated, i.e. the principle of thinning a homogeneous Poisson process with a higher intensity value.
Also note that, for efficiency reasons, the upper bound $\lambda^*$ can be updated even in the case of a rejected inter-arrival time, given the strict monotonicity of $\lambda(t)$ in between event times.
The temporal complexity of sampling $N$ events is $O(N^2)$, since brute-force computation of event intensity using Eq~\eqref{first-lambda} is $O(N)$. 
Furthermore, if event rates decay fast, then the number of rejected samples can be high before there is an accepted new event time. 


%

\begin{algorithm}[tbp]
	\caption{Simulation by thinning.
	}
	\begin{enumerate}[nosep]
	\item Given Hawkes process as in Eq~(\ref{first-lambda}) 
		\item
		Set current time $T = 0$ and event counter $i = 1$
		
		\item
		While $i \leq N$
		\begin{enumerate}[nosep]
			\item Set the upper bound of Poisson intensity $\lambda^*=\lambda(T)$ (using Eq~\eqref{first-lambda}). \label{step:step-3a}
			\item Sample inter-arrival time: draw $u \sim U(0, 1)$ and let $\tau = -\frac{ln(u)}{\lambda^*}$ (as described in Eq~\eqref{eq:sample_poisson}).
			\item Update current time: $T = T + \tau$. \label{step:step-3c}
			
			\item Draw $s \sim U(0, 1)$.
			
			\item
			If $s \leq \frac{\lambda(T)}{\lambda^*}$, accept the current sample: let $T_i = T$ and $i = i + 1$. \\
			Otherwise reject the sample, return to step (a). \label{step:step-3e}
			
		\end{enumerate}
	\end{enumerate}
	\label{alg:simulation-thinning}
\end{algorithm}

\subsection{Efficient sampling by decomposition}

We now outline a more efficient sampling algorithm for Hawkes processes with an exponential kernel that does not resort to rejection sampling. 
Recently proposed by \citet{biao-ecp}, it scales linearly to the number of events drawn. 

First, the proposed algorithm applies to a Hawkes process with exponential immigrant rates and exponential memory kernel. This is a more general form than what we defined in Sec~\ref{ssec:intensity_func}. The  immigrant rate is described by a non-homogenous Poisson process following a exponential function $ a + (\lambda_0 - a) e^{-\delta t}$. For each new event, the {\em jump} it introduces in event intensity is described by a constant $\gamma$. 
\begin{align}
\lambda(t) = a + (\lambda_0 - a) e^{-\delta t} + \sum_{T_i < t} \gamma e^{-\delta (t-T_i)},~t>0 
\label{eq:exp_intensity}
\end{align}
We can envision to generalize this even more by introducing a distribution to $gamma$, this is out of scope for this tutorial.

We note that a process is a Markov process, if it has the property that, conditional on the present, the future is independent of the past. \citet{Ogata1981} showed that the intensity process is a Markov process when $\phi$ is exponential. This can be intuitively understood for event intensity function above, due to $\lambda(t_2) = e^{-\delta(t_2-t_1)} \lambda(t_1)$, for any $t_2 > t_1$. In other words, given current event intensity $\lambda(t_1)$, future intensity only depend on the time elapsed since time $t_1$.

We use this Markov property to decompose the inter-arrival times into two independent simpler random variables. 
The first random variable $s_0$, represents the inter-arrival time of the next event, if it were to come from the constant background rate $a$. It is easy to see that this is sampled according to Eq~(\ref{eq:sample_poisson}). 
The second random variable $s_1$, represents the inter-arrival time of the next event if it were to come from either the exponential immigrant kernel $(\lambda_0 -a) e^{-\delta t}$ or the Hawkes self-exciting kernels from each of the past events $\sum_{T_i < t} e^{-\delta (t-T_i)}$. The cumulative distribution function of $s_1$ can be explicitly inverted due to its Markov property, a full derivation can be found in \citep{biao-ecp}. 
Intuitively the sampled inter-arrival time is the minimum of these two cases. 
It is also worth noting that the second arrival time may not be finite, this is expected, as the exponential kernel decays fast. In this case, the next event will be an immigrant from the constant rate. 
This algorithm is outlined in Algorithm~\ref{alg:simulation-stochastic-hawkes}.
		
\begin{algorithm}[tbp]
	\caption{Simulation of Hawkes with Exponential Kernel}
	\begin{enumerate}[nosep]
		\item
		Set $T_0 = 0$, initial event rate $\lambda(T_0) = \lambda_0$. 
		
		\item
		For $i = 1,2,...,N$
		\begin{enumerate}[nosep]
			\item
			Draw $u_0\sim U(0,1)$ and set $s_0 = - \frac{1}{a} \ln u_0$.
			
			\item
			Draw $u_1 \sim U(0,1)$. Set $d = 1+ \frac{\delta \ln u_1}{\lambda(T_{i-1}^+) - a}$. 
			\item If $d > 0$, set $s_1=-\frac{1}{\delta}\ln d$, $\tau_i = \min\{s_0, s_1\}$.\\
			Otherwise ~$\tau_i = s_0$
			\item Record the $i^{th}$ jump time $T_i = T_{i-1} +  \tau_i$.
			\item Update event intensity at the left side of $T_i$ with exponential decay:\\
			$\lambda(T_{i}^-) = (\lambda(T_{i-1}^+)-a ) e^{-\delta \tau_i} + a$ \label{step:updateT_}
			\item Update event intensity at the right side of $T_i$ with a jump from the $i^{th}$ event:\\
			$\lambda(T_{i}^+) = \lambda(T_{i}^-) + \gamma$ \label{step:updateT+}
		\end{enumerate}
	\end{enumerate}
	\label{alg:simulation-stochastic-hawkes}
\end{algorithm}

This algorithm is efficient because the intensity function can be updated in constant time for each event with steps~\eqref{step:updateT_} and \eqref{step:updateT+}, and that this algorithm does not rely on rejection sampling. 
The decomposition method above cannot be easily used on the power law kernel, since the power law does not have the Markov property. 


\section{Estimation of Hawkes processes parameters} 
\label{sec:fitting-data}
	
One challenge when modeling using self-exciting point processes is estimating parameters from  observed data. 
In the case of the Hawkes process with exponential kernel, one would typically have to determine the function $\lambda_0(t)$ (the base intensity defined in Eq.~\ref{first-lambda}), and the values of the parameters of the decaying kernel $\phi(t)$ ($\alpha$ and $\delta$, see Eq.~\ref{eq:exp_intensity}). 
One can achieve this is by maximizing the likelihood over the observed data. 
In Sec.~\ref{subsec:likelihood} we derive the formula of the likelihood function for a Hawkes process and in Sec.~\ref{subsec:maximum-likelihood-estimation} we discuss a few practical concerns 
of using maximum likelihood estimation. 

\subsection{Likelihood function for Hawkes process}
\label{subsec:likelihood}

Let $N(t)$ be a point process on $[0,T]$ for $T<\infty$ and let $\{T_1,T_2,...,T_n\}$ denote a realization, i.e. the set of event times, of $N(t)$ over the period $[0,T]$. 
Then the data likelihood $L$ as a function of parameter set $\theta$ is:
		\begin{align} \label{eq:max-likelihood}
		L(\theta) = \prod_{i=1}^{n}\lambda(T_i)e^{-\int_{0}^{T}\lambda(t)dt}.
		\end{align}
	
We sketch the derivation of the likelihood formula, along the lines of \citep{DaleyBook,laub2015hawkes,rasmussen-bayesian}. 
If we are currently at some time $t$, recall that the history $\mathcal{H}_t$ is the list of times of events $T_1,T_2,....,T_n$  up to but not including time $t$. Borrowing the $\ast$ notation from \cite{DaleyBook}, we define $f^{\ast}(t):=f(t|\mathcal{H}_t)$ be the conditional probability density function of the time of the next event $T_{n+1}$ given the history of previous event $T_1,T_2,...,T_n$. Recall that $\p\{T_{n+1}\in (t,t+dt)\}=f_{T_{n+1}}(t)dt$. We have
	\begin{align}
	f(T_1,T_2,...,T_n)=\prod_{i=1}^{n}f(T_i|T_1,T_2,...,T_{i-1})=\prod_{i=1}^{n}f^{\ast}(T_i)
	\end{align}

\noindent	
It turns out that the event intensity $\lambda(t)$ can be expressed in terms of the conditional density $f^{\ast}$ and its corresponding cumulative distribution function $F^{\ast}$~\citep{rasmussen2011note}. 
\begin{equation}\label{eq:hazard}
	\lambda(t) = \frac{f^{\ast}(t)}{1-F^{\ast}(t)} \enspace.
\end{equation}
The expression above is given without a formal proof, but it can be interpreted heuristically as follows. Consider an infinitesimal interval $dt$ around t, $f^{\ast}(t)dt$ correspond to the probability that there is an even in $dt$, and $1-F^{\ast}(t)$ correspond to the probability of no new events before time $t$. After manipulating the expression using Bayes rule~\citep{rasmussen2011note}, the ratio of the two can be shown to be equivalent to the expectation of an increment of the counting process $N_{t+dt} - N_t$, which by Eq~\eqref{eq:non-homog-poisson} is essentially $\lambda(t)dt$. 

We can write the conditional intensity function in terms of the cumulative distribution function $F^{\ast}$:
\begin{align}
	\lambda(t)=\frac{f^{\ast}(t)}{1-F^{\ast}(t)}=\frac{\frac{\partial}{\partial t}F^{\ast}(t)}{1-F^{\ast}(t)}=-\frac{\partial}{\partial t}\log(1-F^{\ast}(t)).
\end{align}
Denote the last known event time before $t$ as $T_n$, integrating both sides from $(T_n,t)$, we get
\begin{align}
	\int_{T_n}^{t}\lambda(s)ds = -[\log(1-F^{\ast}(t))-\log(1-F^{\ast}(T_n))].
\end{align}
Note that $F^{\ast}(T_n)=0$ since $T_{n+1}>T_n$ and so
\begin{align}
\int_{T_n}^{t}\lambda(s)ds = -\log(1-F^{\ast}(t)).
\end{align}
Rearranging gives the following expression
\begin{align}
F^{\ast}(t)=1-\exp\left(-\int_{T_n}^t\lambda(s)ds\right)
\end{align}
Combining the relationship between $\lambda(t)$, $f^\ast(t)$, and $F^\ast(t)$ in Eq~\ref{eq:hazard} gives
\begin{align}\label{eq:fstar}
f^\ast(t) = \lambda(t)\left(1-F^{\ast}(t)\right) = \lambda(t) \exp\left(-\int_{T_n}^t\lambda(s)ds\right).
\end{align}
Plugging in Eq~\eqref{eq:fstar} above into the likelihood function, and combining integration ranges, we get the likelihood expression. 
\begin{align}
	L(\theta) =\prod_{i=1}^{n}f^{\ast}(T_i)=
\prod_{i=1}^{n}\lambda(T_i)e^{-\int_{T_{i-1}}^{T_i}\lambda(u)du}=\prod_{i=1}^{n}\lambda(T_i)e^{-\int_0^{T_n}\lambda(u)du}.
\end{align}

\subsection{Maximum likelihood estimation}
\label{subsec:maximum-likelihood-estimation}

Let $\theta$ be the set of parameters of the Hawkes process, its maximum likelihood estimate can be found by maximizing the likelihood function in Eq.~\ref{eq:max-likelihood} with respect to $\theta$ over the space of parameter $\Theta$. 
More precisely, the maximum likelihood estimate $\hat\theta$ is defined to be $\hat{\theta}=\arg\max_{\theta\in\Theta} l(\theta)$. From a standpoint of computational and numerical complexity, we note that summing is less expensive than multiplication. But more importantly, likelihoods would become very small and would risk the running out of floating point precision very quickly, yielding an underflow, thus it is customary to maximize the log of the likelihood function:
\begin{align}
	\label{eq:loglik}
	l(\theta)~= \log L(\theta) ~=-\int_{0}^{T}\lambda(t)dt + \sum_{i=1}^{N(T)}\log \lambda(T_i)
\end{align}
The natural logarithm is a monotonic function and maximizing the log-likelihood automatically implies maximizing the likelihood function. The negative log-likelihood can be minimized with optimization packages for non-linear objective,  such as the L-BFGS~\citep{zhu1997algorithm} software. 

\textbf{Local maxima.}
One may run into problems of multiple local maxima in the log-likelihood. 
The shape of the negative log-likelihood function can be fairly complex and may not be globally convex. 
Due to the possible non-convex nature of the log-likelihood, performing maximum likelihood estimation would result in the estimate being the local maximum rather than the global maximum. 
A usual approach used in trying to identify the global maximum involves using several sets of different initial values for the maximum likelihood estimation. 
Note that this does not mitigate the problem entirely and it is well possible that a local maximum may still be wrongly established as the global maximum.
Alternatively, one can use different optimization methods in conjunction with several different sets of initial values.
If the differing optimizations result in a consistent set of calibrated parameters, then we can have a higher certainty that the calibrated point is the actual global maximum.

\textbf{Edge effects.}
Recall that $N_t$ is the number of `arrivals' or `events' of the process by time $t$ and that the sequence of event times $T_1,T_2,...,T_{N_T}$ is assumed to be observed within the time interval $[0,T]$, where $T<\infty$. 
As discussed in Sec.~\ref{subsec:branching-structure}, in a Hawkes process, the events usually arrive clustered in time: an immigrant and its offspring. 
In practical applications, the process might have started sometime in the past, prior to the moment when we start observing it, denoted as $t = 0$. 
Hence, there may be unobserved event times which occurred before time $0$, which could have generated offspring events during the interval $[0,T]$. 
It is possible that the unobserved event times could have had an impact during the observation period, i.e. sometime after $t>0$, but because we are not aware of them, their contribution to the event intensity is not recorded. 
Such phenomenon are referred to as \textit{edge effects} and are discussed in \citet{DaleyBook} and \citet{rasmussen-bayesian}. 
One possible avenue to address this issue is to assume that the initial value of the intensity process equals the base intensity and disregard edge effects from event times occurring before the observation period, see \citet{DaleyBook}. 
This is usually the modeling setup in most applications within the Hawkes literature. 
As pointed out by \citet{rasmussen-bayesian}, the edge effects on the estimated model would turn out to be negligible if the used dataset is large enough. 
In this chapter, we set the base intensity to be a constant $\lambda(0)=\lambda_0$ and ignore edge effects from events that have occurred before the start of the observation period.
For detailed discussions on handling edge effects, we refer the reader to the extensive works of \citet{MollerRasmussen2005,rasmussen-bayesian,BebbingtonHarte2001,BaddeleyTurner2000,DaleyBook} which are summarized in \citet{lapham} and references therein.

\textbf{Computational bottleneck.}
A major issue with maximum likelihood estimation for Hawkes is the computational costs for evaluating the log-likelihood, in particular the evaluation of the intensity function, as shown here-after. 
Note that the two components of the log-likelihood in Eq.~\eqref{eq:loglik} can be maximized separately since if they do not have common terms, see \cite{Ogata1988,DaleyBook,Zipkin2016}. 
The computational complexity arises due to the calculation of a double summation operation. 
This double sum comes from the second part of the log-likelihood:
\begin{align}
	\label{loglik2}
	\sum_{i=1}^{N_T}\log \lambda(T_i)=\sum_{i=1}^{N_T}\left(\log (a+(\lambda_0-a)e^{-\delta t} + \sum_{j:T_j<T_i}\alpha e^{-\delta(T_i-T_j)})\right).
\end{align}
Note the complexity for most Hawkes process is usually of the order $\mathcal{O}(N_T^2)$, where $N_T$ is the number of event times. 
Hence estimating the parameters can be relatively slow when $N_T$ is of a big number, and it may be exacerbated if loop calculations cannot be avoided. 
In the case of an exponential kernel function, the number of operations required to evaluate Eq.~\eqref{loglik2} can be reduced to $\mathcal{O}(N_T)$ using a recursive formula \citep{Ogata1981}. 
For a more complicated Hawkes process involving a power-law decay kernel, such as the Epidemic Type Aftershock-Sequences (ETAS) model \citep{Ogata1988} or the social media kernel constructed in Sec.~\ref{sec:hawkes-social-media}, this strategy does not hold.
For the ETAS model, the event intensity is defined as:
\begin{align}
	\lambda(t) = \lambda_0 + \sum_{i:t>T_i}\alpha\frac{e^{\delta \eta_1}}{(t-T_i+\gamma)^{\eta_2+1}}
\end{align}
for some constants $\lambda_0,\alpha,\eta_1,\gamma,\eta_2$.
The ETAS model is a point process used typically to represent the temporal activity of earthquakes for a certain geophysical region. 
To reduce the computational complexity for the ETAS model, \cite{Ogata1993} presented a methodology which involved multiple transformations and numerical integration.
They showed that there is a reduction in the time taken to learn the parameters and further demonstrated that they are, in fact, a close approximation of the maximum likelihood estimates.




\section{Constructing a Hawkes model for Social Media}
\label{sec:hawkes-social-media}

The previous sections of this chapter introduced the theoretical bases for working with Hawkes processes.
Sec.~\ref{sec:point-poisson-processes} and~\ref{sec:hawkes-processes} gave the definitions and the basic properties of point processes, Poisson processes and Hawkes processes.
Sec.~\ref{sec:simulation} and~\ref{sec:fitting-data} respectively presented methods for simulating events in a Hawkes process and fitting the parameters of a Hawkes process to data.
The aim of this section is to provide a guided tour for using Hawkes processes with social media data. 
We will start from customizing the memory kernel with a goal of predicting the popularity of an item.
The core techniques here is from a recent paper~\citep{Mishra2016} on predicting the size of a retweet cascade. 
In Sec.~\ref{subsec:construct-kernel} we argue why a Hawkes process is suitable for modeling the retweet cascades and we present the construction of the kernel function $\phi(t)$;
in Sec.~\ref{ssec:loglikelihood} we estimate model parameters from real-life data using Twitter data; 
in Sec.~\ref{subsec:final-retweets-estimation} we predict the expected size of a retweet cascade, i.e. its popularity.


\subsection{A marked Hawkes process for information diffusion}
\label{subsec:construct-kernel}

We model \emph{word of mouth} diffusion of online information: users share content, and other users consume and sometimes re-share it, broadcasting to more users.
For this application, we consider each retweet as an event in the point process. We also formulate information diffusion in Twitter as a self-exciting point process, in which we model three key intuitions of the social network:
\emph{magnitude of influence}, tweets by users with many followers tend to get retweeted more; 
\emph{memory over time}, that most retweeting happens when the content is {\em fresh}~\cite{Wu2007};
and \emph{content quality}.

\textbf{The event intensity function.}
A retweet is defined as the resharing of another person's tweet via the dedicated functionality on the Twitter interface. A retweet cascade is defined as the set of retweets of an initial tweet.
Using the branching structure terminology introduced in Sec.~\ref{sec:hawkes-processes}, a retweet cascade is made of an \emph{immigrant} event and all of its \emph{offsprings}.
We recall the definition of the event intensity function in a Hawkes process, introduced in Eq.~\eqref{first-lambda}:
\begin{equation} \label{eq:hawkes}
	\lambda(t) = \lambda_0(t) + \sum_{T_i < t} \phi_{m_i}(t - T_i)\enspace.
\end{equation}
$\lambda_0(t)$ is the arrival rate of immigrants events into the system.
The original tweet is the only immigrant event in a cascade, therefore $\lambda_0(t) = 0, \forall t > 0$.
Furthermore, this is modeled as a {\em marked} Hawkes process. The {\em mark} or magnitude of each event models the user influence for each tweet.
The initial tweet has event time $T_0=0$ and mark $m_0$. Each subsequent tweet has the mark $m_i$ at event time $T_i$.

We construct a power-law kernel $\phi_m(\tau)$ with mark $m$:
\begin{equation} \label{eq:phi-separable}
	\phi_m(\tau) = \kappa m^{\beta} (\tau + c)^{-(1+\theta)} \enspace.
\end{equation}
$\kappa$ describes the {\em virality} -- or quality -- of the tweet content and it scales the subsequent retweet rate; 
$\beta$ introduces a warping effect for user influences in social networks;
and $1+\theta$ ($\theta>0$) is the power-law exponent, describing how fast an event is {\em forgotten}, parameter $c>0$ is a temporal shift term to keep $\phi_m (\tau)$ bounded when $\tau \simeq 0$.
Overall, $\kappa m^{\beta}$ accounts for the magnitude of influence, and the power-law kernel $(\tau + c)^{-(1+\theta)}$ models the memory over time.
We assume user influence $m$ is observed the number of followers obtained from Twitter API.

In a similar fashion, we can construct an exponential kernel for social media, based on the kernel defined in Eq.~\eqref{eq:exp-kernel}:
\begin{equation} \label{eq:exp-kernel-social-media}
	\phi_m(\tau) = \kappa m^{\beta} \theta e^{- \theta \tau} \enspace.
\end{equation}
We have experimented with this kernel and Fig.~\ref{fig:point-process-interpretation}(c) shows the its corresponding intensity function over time for a real twitter diffusion cascade.
However, we have found that the exponential kernel for social media provides lower prediction performances compared to the power-law kernel defined in Eq.~\ref{eq:phi-separable}.
Consequently, in the rest of this chapter, we only present the power-law kernel.

\subsection{Estimating the Hawkes process}
\label{ssec:loglikelihood}

The marked Hawkes process has four parameters $\theta = \{ \kappa, \beta, c, \theta\}$, which 
we set out to estimate using maximum likelihood estimation technique described in Sec.~\ref{sec:fitting-data}.
We can obtain the its log-likelihood by introducing the marked memory kernel \eqref{eq:phi-separable} into the general log-likelihood formula shown in Eq.~\eqref{eq:loglik}. The first two terms in Eq.~\ref{eq:ll} are from the likelihood computed using the event rate $\lambda(t)$, the last term is a normalization factor from integrating the event rate over the observation window $[0,T]$. 
\begin{align}
{\cal L}(\kappa, \beta, c, \theta) =& \sum_{i=2}^n\log\kappa + \sum_{i=2}^n \log\left( \sum_{t_j<t_i}\dfrac{\left(m_j\right)^\beta}{\left(t_i - t_j + c\right)^{1+\theta}}\right) \nonumber \\ 
&- \kappa \sum_{i=1}^n \left({m_i}\right)^\beta \left[\dfrac{{1}}{\theta c^{\theta}} - \dfrac{\left(T+c-t_i\right)^{-\theta}}{\theta} \right] \enspace. \label{eq:ll}
\end{align}

Eq.~\ref{eq:ll} is a non-linear objective that need to be maximized. 
There are a few natural constraints for each of model parameter, namely: $\theta>0$, $\kappa>0$, $c>0$, and $0<\beta<\alpha-1$ for the branching factor to be meaningful (and positive).
Furthermore, while the supercritical regimes $n^{\ast} > 1$ are mathematically valid, it will lead to a prediction of infinite cascade size -- a clearly unrealistic outcome. 
We further incorporate $n^{\ast} < 1$ as a non-linear constraint for the maximum likelihood estimation.
\texttt{Ipopt}~\cite{Wachter2006}, the large-scale interior point solver can be used to handles both non-linear objectives and non-linear constraints.
For efficiency and precision, it needs to be supplied with pre-programmed gradient functions. 
Details of the gradient computation and optimization can be found in the online supplement \citep{Mishra2016}.

Sec.~\ref{subsec:maximum-likelihood-estimation} warned about three possible problems that can arise when using maximum likelihood estimates with Hawkes processes: edge effects, squared computational complexity and local minima.
In this application, since we always observe a cluster of events generated by an immigrant, we do not having \emph{edge effects}, i.e, missing events early in time.
The \emph{computational complexity} of calculating the log-likelihood and its gradients is $O(n^2)$, or quadratic with respective to the number of observed events.
In practice, we use three techniques to make computation more efficient: vectorization in the R programming language, storing and reusing parts of the calculation, and data-parallel execution across a large number of cascades. With these techniques, we estimated tens of thousands of moderately-sized retweet cascades containing hundreds of events in reasonable amount of time. 
Lastly, the problem of \emph{local minima} can be addressed using multiple random initializations, as discussed in Sec.~\ref{subsec:maximum-likelihood-estimation}.

\subsection{The expected number of future events}
\label{subsec:final-retweets-estimation}

Having observed a retweet cascade until time $T$ for a given Hawkes process, one can simulate a possible continuation of the cascade using the thinning technique presented in Sec.~\ref{sec:simulation}.
Assuming a subcritical regime, i.e. $n^{\ast} < 1$, the cascade is expected to die out in all possible continuation scenarios.
In addition to simulation a handful of possible endings, it turns out there is a close-form solution to 
the expected number of future events in the cascade over all possible continuations, 
i.e., the total popularity that the cascade will reach at the time of its ending.

There are three key ideas for computing the expected number of future events. The first is to compute the expected size of a direct offsprings to a event at $T_i$ after time $T$; the second is that the expected number all descendent events can be obtained via the branching factor of the Hawkes process, as explained in Sec.~\ref{subsec:branching-structure}. Lastly, the estimate of total popularity emerges when we put these two ideas together. 

\textbf{The number of future children events.}
In retweet cascades, the base intensity is null $\lambda_0(t) = 0$, therefore no new immigrants will occur at $t > T$.
Eq.~\eqref{eq:n-inf-cluster-sumgp} gives the expected size of a cluster of offprings associated with an immigrant.
In the marked Hawkes process Eq~\eqref{eq:hawkes}, each of the $i=1,\dots,n$ events that happened at $T_i<T$ adds $\phi_{m_i}(t-T_i)$ to the overall event intensity. 
We can obtain the expectation of $A_1$, the total number of events directly triggered by event $i=1,\dots,n$, by integrating over the memory kernels of each event. The summation and integration are exchangeable here, since the effect of each event on future event intensity is additive. 
\begin{align} \label{eq:A1}
	A_1 &= \int_T^{\infty} \lambda(t)\mathrm{d} t = \int_T^{\infty} \sum_{t>T_i} \phi_{m_i}(t-T_i) \mathrm{d}t \nonumber \\ 
	&=  \sum_{t>T_i} \int_T^{\infty} \phi_{m_i}(t-T_i ) \mathrm{d}t  
	= {\kappa} \sum_{i=1}^n \dfrac{{m_i}^{\beta}}{\theta \left(T+c-T_i\right)^{\theta}}
\end{align}
%
%

\textbf{The branching factor}
The branching factor was defined in Eq.~\eqref{eq:nstar-general} for an unmarked Hawkes process.
We compute the branching factor of the marked Hawkes process constructed in Sec~\ref{subsec:construct-kernel} by taking expectations over both event times and event marks.
We assume that the event marks $m_i$ are {\em i.i.d.} samples from a power law distribution of social influence~\citep{kwak2010twitter}: $P(m) = (\alpha - 1) \: m^{-\alpha}$.
$\alpha$ is an exponent which controls the heavy tail of the distribution and it is estimated from a large sample of tweets.
We obtain the closed-form expression of the branching factor (see~\citet{Mishra2016} for details):
\begin{equation} \label{eq:nstar}
	n^{\ast} = \kappa \frac{\alpha - 1}{\alpha - \beta - 1} \frac{1}{\theta c^\theta}, \text{  for } \beta < \alpha - 1 \text{ and } \theta > 0 \enspace.
\end{equation}

\textbf{Total size of cascade.}
Putting both Eq~\eqref{eq:A1} and Eq~\eqref{eq:nstar} together, we can see that each expected event in $A_1$ 
is expected to generate $n^\ast$ direct children events, $n^{\ast 2}$ grand-children events, $\ldots$, 
$n^{\ast k}$ k-th generation children events, and so on. 
The calculation of geometric series shows that the number of all descendants is $\frac{A_1}{1-n^*}$. 
This quantity plus the observed number of events $n$ is the total number of expected events in the cascade. 
See~\citep{Mishra2016} for complete calculations. 

\begin{align} \label{eq:Ninf}
N_{\infty} = n + \dfrac{\kappa} {(1-n^\ast)} \left( \sum_{i=1}^n \dfrac{{m_i}^{\beta}}{\theta \left(T+c-t_i\right)^{\theta}}\right), n^\ast < 1
\end{align}

\subsection{Interpreting the generative model}
\label{subsec:model-interpretation}

\begin{figure}[htbp]
	\centering
	\includegraphics[width=0.99\textwidth]{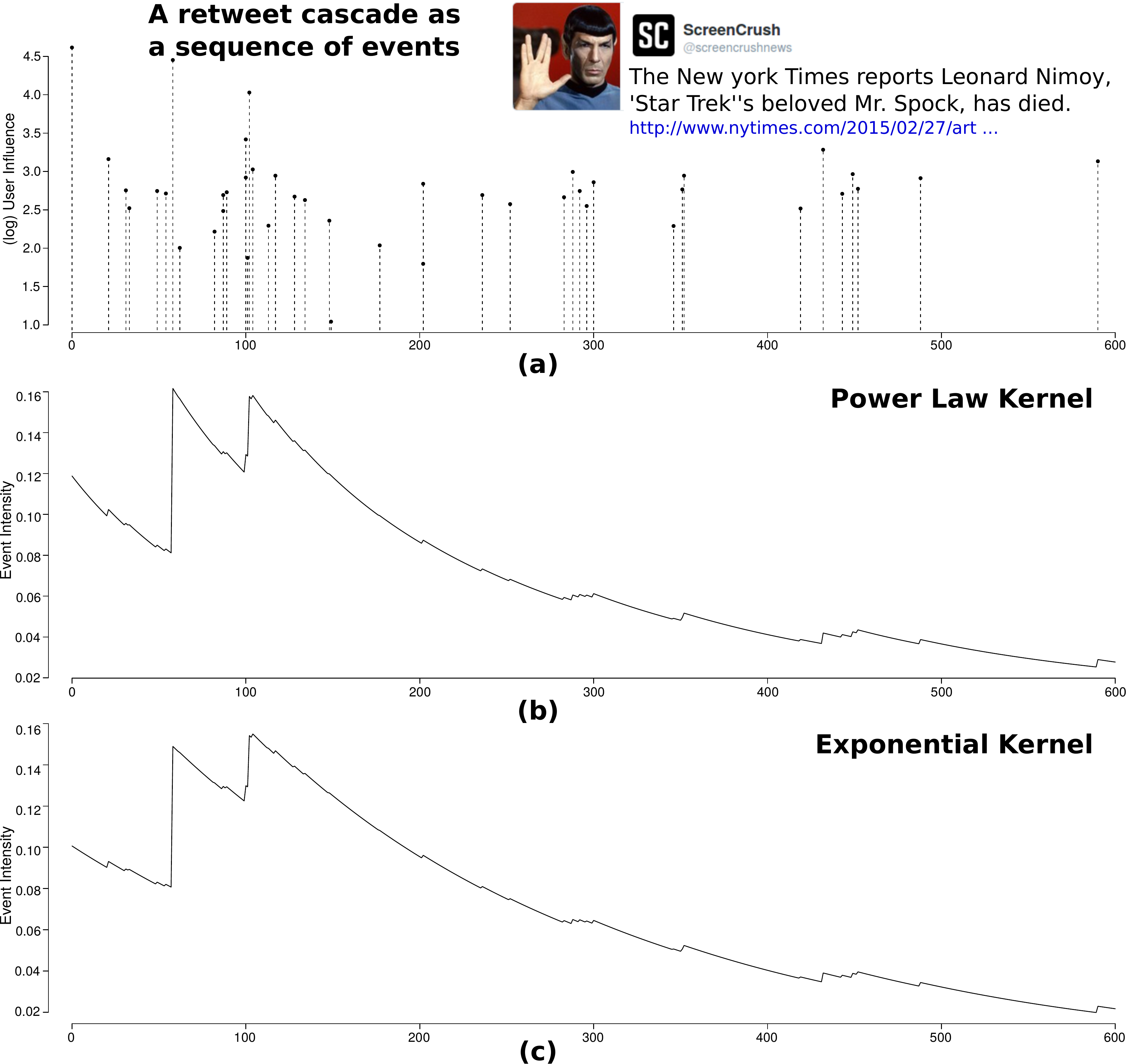}
	\caption{
		An example retweet cascade on a \href{http://www.nytimes.com/2015/02/27/arts/television/leonard-nimoy-spock-of-star-trek-dies-at-83.html}{news article by The New York Times}. 
		(a) Representation of the first 600 seconds of the retweet cascade as a marked point process, to each (re)tweet corresponds an event time.
		(b) Event intensity ($\lambda(t)$) over time, assuming the point process to be a Hawkes process with power-law kernel. 
		The maximum-likelihood  model parameter estimates are $\{\kappa = 1.00, \beta = 1.01, c = 250.65, \theta =  1.33 \}$ with a corresponding $n^\ast = 0.92$ and a predicted cascade size of $216$. 
		The true cascade size is $219$.
		(c) The event intensity over time for the same event time series, when the point process is assumed to the a Hawkes process with the exponential kernel define din Eq.~\ref{eq:exp-kernel-social-media}.
		The fitted parameters for this kernel are $\{\kappa = 0.0003, \beta = 1.0156, \theta =  0.0054 \}$, the corresponding $n^\ast = 0.997$ and the predicted cascade size is $1603$.  
	}
	\label{fig:point-process-interpretation}
\end{figure}

A Hawkes process is a generative model, meaning that it can be used to interpret statistical patterns in diffusion processes, in addition to being used in predictive tasks. 
Fig.~\ref{fig:point-process-interpretation} presents a diffusion cascade about a New York Times news article with its corresponding intensity functions with the power-law and exponential memory kernels, respectively.
Note that the top and lower two graphics are temporally aligned. In other words, each occurred event causes a jump in the intensity function, i.e. increasing the likelihood of future events.
Each jump is followed by a rapid decay, governed by the decay kernel $\phi_m(\tau)$, defined in Sec~\ref{subsec:construct-kernel}.
In terms of event marks, the cascade attracts the attention of some very well-followed accounts.
The original poster (\texttt{@screencrushnews}) has 12,122 followers, and among the users who retweeted,  \texttt{@TasteOfCountry} (country music) has 193,081 followers, \texttt{@Loudwire} (rock) had 110,824 followers, \texttt{@UltClassicRock} (classic rock) has 99,074 followers and \texttt{@PopCrush} (pop music) has 114,050 followers. 

For popularity prediction, the cascade is observed for 10 minutes (600 seconds) and the parameters of the Hawkes process are fitted as shown in Sec.~\ref{ssec:loglikelihood}.
The maximum-likelihood estimate of parameters with a power-law kernel are $\{\kappa = 1.00, \beta = 1.01, c = 250.65, \theta =  1.33 \}$, with a corresponding $n^\ast = 0.92$. 
According to the power-law kernel, this news article has high content virality (denoted by $\kappa$) and large waiting time ($c$), which in turn lead to a slow diffusion: the resulting cascade reached 1/4 its size after half an hour, and the final tweet was sent after 4 days.
By contrast, most retweet cascades finish in a matter of minutes, tens of minutes at most.
Using the formula in Eq.~\eqref{eq:Ninf}, we predict the expected total cascade size $N_\infty = 216$, this is very close to the real cascade size of $219$ tweets, after observing only the initial 10 minutes of the 4 day Twitter diffusion.
When estimated with an exponential kernel, the parameters of Hawkes point process are $\{\kappa = 0.0003, \beta = 1.0156, \theta =  0.0054 \}$ and the corresponding branching factor is $n^\ast = 0.997$. 
This produces an very imprecise total cascade size prediction of $1603$ tweets, largely due to the high $n^\ast$.


%

\subsection{Hands-on tutorial}
\label{ssec:handson}

In this section, we provide a short hand-on tutorial, together with code snippets required for modeling information diffusion through retweet cascades. 
A detailed version of tutorial with example data and code is available at \url{https://github.com/s-mishra/featuredriven-hawkes}. 
All code examples presented in this section assume a Hawkes model with the power-law kernel.
The complete online tutorial also presents examples which use an exponential kernel.
All code was developed using the R programming language.

\begin{figure}[bp]
	\centering
	\includegraphics[width=0.95\textwidth]{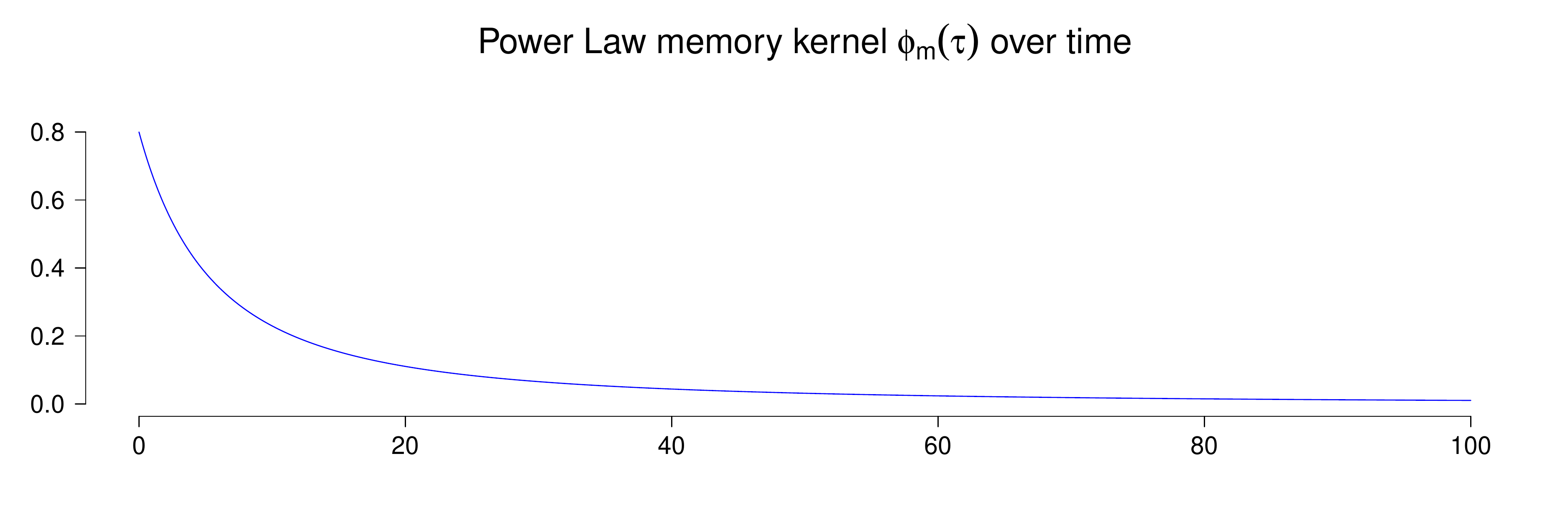}
	\caption{
		Graphic obtained by running the code in Listing~\ref{code:PLKernel}:
		The power-law kernel over time, generated by an event with the mark 1000.
	}
	\label{fig:PLKernel}
\end{figure}

We start with visualizing in Fig.~\ref{fig:PLKernel} the shape of the power-law kernel (defined in Eq.~\eqref{eq:phi-separable}) generated by an event with the mark $m = 1000$, and defined by the parameters $\kappa = 0.8$, $\beta = 0.6$, $c = 10$ and $\theta = 0.8$.
The code for generating the figure is shown in Listing.~\ref{code:PLKernel}.
Furthermore, we can simulate (Listing~\ref{code:simulation}) the entire cluster of offspring generated by this initial immigrant event using the thinning procedure described in Sec.~\ref{subsec:thinning-simulation}.
The initial event is assumed to have occurred at time $t = 0$, and the simulation is ran for 50 time intervals.

We now show to estimate the parameters of a Hawkes process with a power-law kernel for a real Twitter diffusion cascade and how to estimate the total size of the cascade.
The file \texttt{example\_book.csv} in the online tutorial records the retweet diffusion cascade around a news article announcing the death of ``Mr. Spock'' shown in Fig.~\ref{fig:point-process-interpretation}.
Fig.~\ref{fig:point-process-interpretation}(a) depicts the cascade as a point process: 
the tweet posting times are the event times, whereas the number of followers of the user emitting the tweets are considered the event marks.
The code in Listing~\ref{code:fitting} reads the CSV file and performs a maximum likelihood estimation of the Hawkes process parameters, based on the events in the cascade having occurred in the first 600 seconds (10 minutes).
With the obtained estimates for model parameters, we can predict (using the code in Listing~\ref{code:prediction}) the total size of the diffusion cascade.

\begin{lstlisting}[language=R,
				caption={Code for computing the power-law kernel function, generated by an event with mark 1000.},
				label=code:PLKernel, 
				float=b]
## initial event that starts the cascade (i.e. the immigrant)
## mark = 1000 at time t = 0
event <- c(mark = 1000, time = 0)
## the timepoints for which we compute the kernel function
t <- seq(from = 0, to = 100, length=1000)
## set the parameters of the kernel
K <- 0.8
beta <- 0.6
c <- 10
theta <- 0.8

## compute the Power Law Kernel
## call the kernelFunction to get the values
values.PL <- kernelFct(event = event, t = t, K = K, beta = beta, c = c,
                        theta = theta, kernel.type='PL')

## plot the obtained kernel
plot(x = t, y = values.PL, type = "l", col = "blue", 
     xlab = "", ylab = "", main = "Power-law memory kernel over time")
\end{lstlisting}

\begin{lstlisting}[language=R,
				caption={Simulation using the thinning method (see Sec.~\ref{subsec:thinning-simulation}) of an entire cluster of offspring, generated by the immigrant defined in Listing~\ref{code:PLKernel} ($m = 1000, t = 0$).},
				label=code:simulation, 
				float=t]
## simulating an event series
events <- generate_Hawkes_event_series(K = K, beta = beta, c = c, theta = theta, 
                                       M = event["mark"], Tmax = 50)
\end{lstlisting}

\begin{lstlisting}[language=R,
				caption={Load the information about the real tweet cascade from Fig.~\ref{fig:point-process-interpretation}, and fit parameters using the events observed in the first 600 seconds.},
				label=code:fitting, 
				float=t]
## read the real cascade provided in the file "example.csv"
real_cascade <- read.csv(file = 'example_book.csv', header = T)
## retain only the events that occurred in the first 600 seconds (10min). These
## will be used for fitting the model parameters.
predTime <- 600
history <- real_cascade[real_cascade$time <= predTime, ]
## removing the first column, which  is event index
## retaining column 2 (event mark) and column 3 (event time).
history <- history[ , 2:3]
## call the fitting function, which uses IPOPT internally. The fitting algorithm
## requires an initial guess of the parameters. This can a random point within
## the domain of definition of parameters.
startParams <- c(K = 1, beta = 1, c = 250, theta = 1)
result <- fitParameters(startParams, history)
\end{lstlisting}

\begin{lstlisting}[language=R,
				caption={Predict the total size for the twitter cascade shown in Fig.~\ref{fig:point-process-interpretation}, using the parameters fitted as in Listing~\ref{code:fitting}},
				label=code:prediction, 
				float=t]
## Using the fitted model parameters, we call getTotalEvents to get predictions.
prediction <- getTotalEvents(history = history, bigT = predTime,
                             K = result$solution[1], 
                             beta = result$solution[2],
                             c = result$solution[3],
                             theta = result$solution[4])
## The "predction" object contains other values, such as 
## the branching factor (nstor) and A1
nPredicted = prediction['total']
\end{lstlisting}


\section{Summary}

This chapter provided a gentle introduction for Hawkes self-exciting process. 
We covered the key definitions of point processes and Hawkes processes. We introduced the notion of event rate, branching factor, and the use of these quantities to predict future events. We described procedures for simulating a Hawkes process, and derived the likelihood function used for parameter estimation. We also included a practical example for estimating a Hawkes process from retweet cascades, along with code snippets and online notebook. Where applicable, we have  included discussions of the point-process literature. 
The goal of the materials above is to provide the fundamentals to researchers who are interested in formulating and solving application problems with point processes. 
Interested readers are invited to explore more advanced materials, including: alternative inference algorithms such as using expectation-maximization, sampling, or moment matching;
flexible specifications and extensions of self-exciting processes such as 
multi-variate mutually-exciting Hawkes processes, doubly-stochastic processes, to name a few.  


	

\backmatter
\bibliography{biblio-LexingXie}



\end{document}